%% file: main.tex
\documentclass[10pt,twocolumn,letterpaper]{article}

\usepackage[pagenumbers]{iccv} % To force page numbers, e.g. for an arXiv version


\definecolor{iccvblue}{rgb}{0.21,0.49,0.74}
\usepackage[pagebackref,breaklinks,colorlinks,allcolors=iccvblue]{hyperref}

\usepackage{subcaption}
\usepackage{bm}
\usepackage{multirow}
\usepackage{adjustbox}
\usepackage{xspace}
\usepackage{pifont}
\usepackage{wrapfig}
\newcommand{\cmark}{\ding{51}}%
\newcommand{\xmark}{\ding{55}}%
\newcommand{\ieno}{\textit{i}.\textit{e}.}
\usepackage{color, colortbl}
\definecolor{Gray}{gray}{0.9}
\usepackage[accsupp]{axessibility}  % Improves PDF readability for those with disabilities.

\def\paperID{****} % *** Enter the Paper ID here
\def\confName{ICCV}
\def\confYear{2025}

\title{Perceiving and Acting in First-Person: \\A Dataset and Benchmark for Egocentric Human-Object-Human Interactions}
\newcommand*{\affmark}[1][*]{\textsuperscript{#1}}

\author{
Liang Xu\affmark[1,2,3] \quad
Chengqun Yang\affmark[1] \quad
Zili Lin\affmark[1,2,3] \quad
Fei Xu\affmark[1] \quad
Yifan Liu\affmark[1] \quad
Congsheng Xu\affmark[1] \\
Yiyi Zhang\affmark[4] \quad
Jie Qin\affmark[5] \quad
Xingdong Sheng\affmark[6] \quad
Yunhui Liu\affmark[6] \quad
Xin Jin\affmark[2,3] \quad
Yichao Yan\affmark[1*] \\
Wenjun Zeng\affmark[2,3] \quad
Xiaokang Yang\affmark[1]
\vspace{0.7em} \\
\affmark[1]{MoE Key Lab of Artificial Intelligence, AI Institute, Shanghai Jiao Tong University} \\
\affmark[2]{Ningbo Institute of Digital Twin, Eastern Institute of Technology, Ningbo, China} \\
\affmark[3]{Ningbo Key Laboratory of Spatial Intelligence and Digital Derivative, Ningbo, China} \\
\affmark[4]{MoE Key Lab of AI, School of Computer Science, Shanghai Jiao Tong University} \\
\affmark[5]{Nanjing University of Aeronautics and Astronautics} \quad
\affmark[6]{Lenovo}\\
{\small\url{https://liangxuy.github.io/InterVLA/}}
\vspace{-0.5em}
}

\begin{document}

\twocolumn[{
    \renewcommand\twocolumn[1][]{#1}
    \maketitle
    \begin{center}
        % \vspace{-6mm}
        \includegraphics[width=1.0\linewidth]{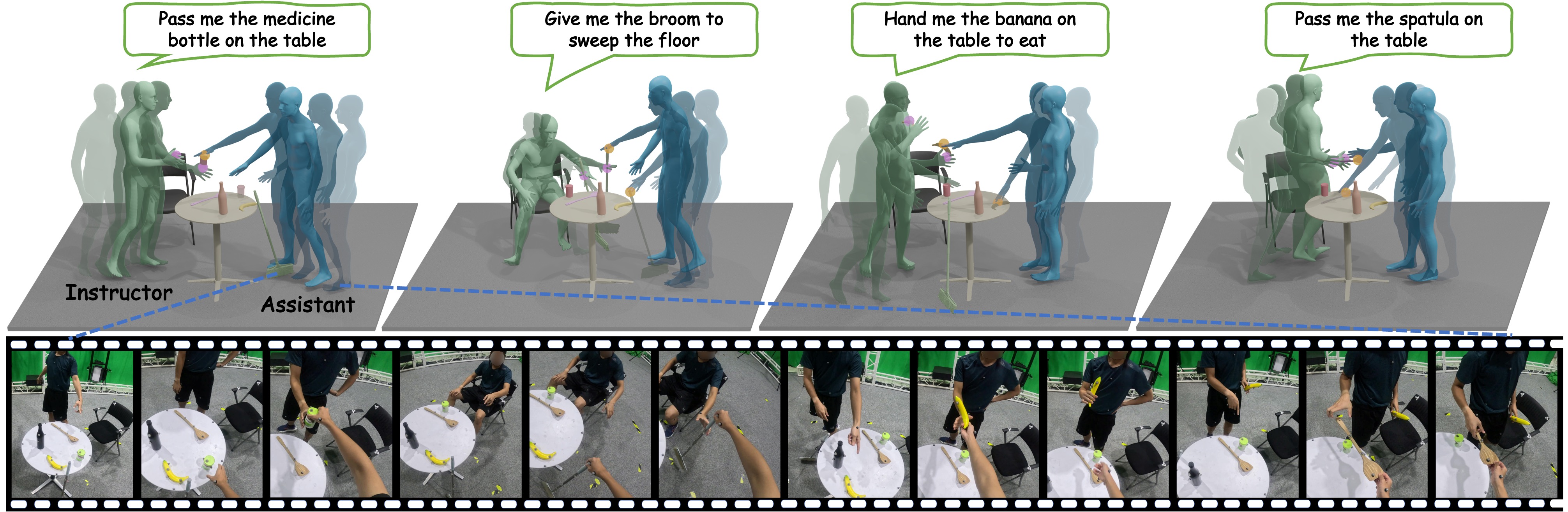}
        % \fbox{\rule{0pt}{2in} \rule{.9\linewidth}{0pt}}
        \captionof{figure}{InterVLA features a large-scale human-object-human interaction dataset in a vision-language-action scheme, where an assistant provides services to an instructor based on egocentric perception and verbal commands. This comprehensive dataset comprises \textbf{3.9K} sequences, totaling \textbf{11.4} hours and \textbf{1.2M} frames of multimodal interaction data, including egocentric and exocentric RGB videos, language commands and high-precision human/object motions, promoting the development of general-purpose intelligent AI assistants.}
        \label{fig:teaser}
        % \vspace{-2mm}
    \end{center}
}]

\let\thefootnote\relax\footnotetext{$^*$Corresponding authors}

\begin{abstract}
Learning action models from real-world human-centric interaction datasets is important towards building general-purpose intelligent assistants with efficiency.
However, most existing datasets only offer specialist interaction category and ignore that AI assistants perceive and act based on first-person acquisition.
We urge that both the generalist interaction knowledge and egocentric modality are indispensable.
In this paper, we embed the manual-assisted task into a vision-language-action framework, where the assistant provides services to the instructor following egocentric vision and commands.
With our hybrid RGB-MoCap system, pairs of assistants and instructors engage with multiple objects and the scene following GPT-generated scripts.
Under this setting, we accomplish InterVLA, the first large-scale human-object-human interaction dataset with 11.4 hours and 1.2M frames of multimodal data, spanning 2 egocentric and 5 exocentric videos, accurate human/object motions and verbal commands.
Furthermore, we establish novel benchmarks on egocentric human motion estimation, interaction synthesis, and interaction prediction with comprehensive analysis. We believe that our InterVLA testbed and the benchmarks will foster future works on building AI agents in the physical world.

\end{abstract}

% \vspace{-4mm}
\section{Introduction}
\label{sec:intro}
% \vspace{-2mm}
% Background
% Significance of hhi and hoi
Learning generalist interaction knowledge is indispensable towards general-purpose intelligent agents to assist humans in the physical world.
To avoid expensive robot data collection, learning from human-centric interactive datasets is more efficient~\cite{gen2act,ma2024diff-ip2d,mandikal2022dexvip,qin2022dexmv}. 
Existing datasets on human-human interactions contribute to human-robot interactions~\cite{jiang2024harmon,mascaro2024hoi4abot,prasad2024moveint,butepage2020imitating}, teleoperation~\cite{seo2023deep,he2024omnih2o,he2024learning}; human-object interactions promote human-to-robot handover~\cite{wang2024genh2r,cini2019choice,wiederhold2024hoh,carfi2019multi}, human-robot collaboration~\cite{zhang2024core4d,sasagawa2020imitation,cunha2020towards} and human-scene interactions advance navigation~\cite{xiao2023unified,zhang2024scenic,zhang2025enhancing}.
% nikolaidis2015efficient

% with versatile open-world tasks

% The significance of egocentric vision, show some examples
% 1. Single-human or single-object, without cooperation; Multiple 
% 2. Egocentric view
% Add existing dataset and **methods** discussions
Despite the rapid development of magnitude and richness in human-centric datasets and benchmarks, they still face some limitations in building intelligent assistants.
Imagining the most basic capabilities for home robots, perceiving and comprehending the instructor's commands, navigating smoothly, and manipulating objects are required. 
However, most datasets only offer specialist interaction category~\cite{ng2020you2me,expi,hi4d,interhuman,inter-x,lv2025himo,wiederhold2024hoh,zhang2024core4d} rather than a generic scenario composed of diverse human-human, object and scene interactions.
Besides, existing datasets~\cite{expi,hi4d,interhuman,inter-x,lv2025himo,wiederhold2024hoh,HOI-M3} ignore the fact that AI assistants always perceive and then react based on their first-person acquisition~\cite{wang2023holoassist,zhang2024core4d}. 
The absence of egocentric perspective could hinder the physical deployment of AI assistants.
% One significant vision of modeling human-human interactions and human-object interactions is to apply the learned capabilities in intelligent robots. However, most of previous works ignore that for intelligent agents, they perceive the world through the egocentric views.

% Our proposal, vla setting
% 1. We want to build a generalist dataset with egocentric videos
% 2. Inspired by the vla setting in other works
% 3. How we plan to build the dataset
% 4. How to build our InterVLA dataset, the vision, language and action part
To address these limitations and foster the development of general human-centric interactions and versatile AI assistants, a comprehensive dataset encompassing \textbf{diverse interaction patterns} and \textbf{stable egocentric perception} is pivotal.
In this paper, we focus on the common daily scenarios of manual-assisted tasks with the majority being human-object-human interactions where an assistant providing services to an instructor following the egocentric vision and verbal commands, such as \textit{``Pass me the cup on the table''}, where the human-human, object, scene interactions are naturally integrated.
To simulate real-world robotic assistance scenarios, we randomly arrange various pieces of furniture and additional operable objects to establish the scene. 
The instructor gives verbal commands accompanied by body gestures while the assistant comprehends the intention and then responds accordingly.

% How to build the dataset, from humans, objects, scenes
% -> RGB-MoCap system, vision-language-action properties
% Properties of the InterVLA dataset
Inspired by the vision-language-action (VLA) paradigm emerged for instruction-following robots, we formulate our data collection setup within the VLA framework and introduce InterVLA, the first large-scale egocentric human-object-human interaction dataset with various interaction categories as depicted in~\cref{fig:teaser}. For the \textit{vision} modality, we capture two egocentric videos from the instructor's perspective and five exocentric videos covering the full scene. The \textit{language} component consists of 100 meticulously crafted scripts featuring various scene arrangements, versatile interaction types, multi-object interactions and navigation tasks. To acquire \textit{action} data, we attach reflective markers to the human and object surfaces, enabling high-precision motion tracking while preserving RGB data fidelity. We recruit \textbf{47} participants to form \textbf{27} unique instructor-assistant pairs engaging with \textbf{50} household objects, yielding \textbf{3.9K} sequences of \textbf{11.4} hours and \textbf{1.2M} frames of interaction data in total. All the captured data are well-calibrated and temporally synchronized.
A comparison with existing human-centric interactive datasets is summarized in~\cref{tab:dataset}.

\begin{table}[t]
  \centering
  \resizebox{0.48\textwidth}{!}{
  \begin{tabular}{@{}lcccccc|ccc@{}}
    \toprule
    Dataset & \multicolumn{6}{c}{\textbf{Modality}}  & \multicolumn{3}{c}{\textbf{Scale}} \\
    & HHI & HOI & HSI & Multi-Obj & Ego & Exo & \#Seqs & \#Objs & \#Hours \\ \hline
    You2Me~\cite{ng2020you2me} & \cmark & \xmark & \xmark & \xmark & \cmark & \xmark & 42 & - & 1.4 \\
    ExPI~\cite{expi} & \cmark & \xmark & \xmark & \xmark & \xmark & \cmark & 115 & - & 0.3 \\
    Hi4D~\cite{hi4d} & \cmark & \xmark & \xmark &  \xmark & \xmark & \cmark & 100 & - & - \\
    InterHuman~\cite{interhuman} & \cmark & \xmark & \xmark & \xmark & \xmark & \cmark & 6.0K & - &  6.6\\
    Inter-X~\cite{inter-x} & \cmark & \xmark & \xmark & \xmark & \xmark & \xmark & 11.4K & - & 18.8 \\
    HIMO~\cite{lv2025himo} & \xmark & \cmark & \xmark & \cmark & \xmark & \xmark & 3.4K & 53 & 9.4 \\
    HOH~\cite{wiederhold2024hoh} & \xmark & \cmark & \xmark & \xmark & \xmark & \cmark & 2.7K & 136 & - \\
    CORE4D~\cite{zhang2024core4d} & \cmark & \cmark & \xmark & \xmark & \cmark & \cmark & 1.0K & 37 & -\\
    HOI-M$^3$~\cite{HOI-M3} & \cmark & \cmark & \cmark & \cmark & \xmark & \cmark & 199 & 90 & 20 \\
    \midrule
    % 3,906 train: 3151, test: 571, val: 184
    InterVLA & \cmark & \cmark & \cmark & \cmark & \cmark & \cmark & 3.9K & 50 & 11.2\\
    \bottomrule
  \end{tabular}}
  \caption{\textbf{Dataset comparison.} We compare InterVLA with existing human-centric interactive datasets. \textbf{Modality} measures the human-human, human-object, human-scene interactions, multi-object interactions, egocentric and exocentric views. \textbf{Scale} measures the number of sequences, objects and hours.}
  \label{tab:dataset}
  % \vspace{-6mm}
\end{table}

\begin{figure*}[t]
  % \vspace{-2mm}
  \centering
  \includegraphics[width=1.0\linewidth]{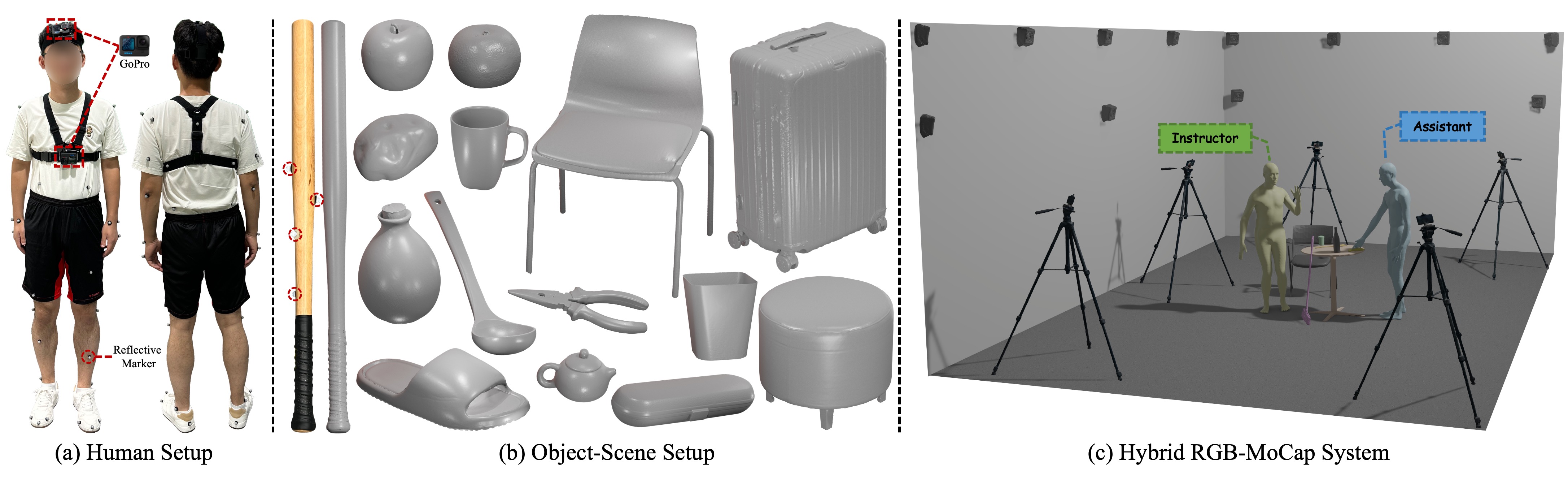}
  % \fbox{\rule{0pt}{2in} \rule{.9\linewidth}{0pt}}
  % \vspace{-2mm}
  \caption{\textbf{InterVLA capturing system.} (a). We attach 41 reflective markers to the subject's body with strong medical glue to track the human body motion. The \textit{assistant} wears two GoPro cameras to capture the egocentric data. (b). Reflective markers are positioned on the surface of real objects to track the precise object trajectories. We also collect the precise 3D object scans with a KSCAN Magic Scanner. (c). Our hybrid RGB-MoCap system with two egocentric RGB cameras, five exocentric RGB cameras, and an OptiTrack MoCap system.}
  \label{fig:data_capture}
  % \vspace{-5mm}
\end{figure*}

% Three benchmarks
% 1. Egocentric human motion estimation
% 2. HOH interaction synthesis
% 4. Interaction prediction
With our proposed InterVLA dataset, we introduce novel tasks and benchmarks on how AI assistants better perceive the surroundings and then generate appropriate responses. We formulate four downstream tasks of 1) Egocentric human motion estimation to extract the global body motion of the instructor based on the first-person perspective of the assistant, 2) Interaction synthesis to generate plausible human-object-human sequences given the textual descriptions and the initial states of the human and objects, 3) Motion-based interaction prediction to anticipate the future human/object motions conditioned on the previous motion frames and 4) Vision-language based interaction prediction to forecast the future human motions from the historical first-person videos and the verbal instruction.
We establish comprehensive benchmarks for these tasks, providing baseline models, quantitative and qualitative evaluations, and in-depth analysis.
The results highlight the challenges posed by rapid camera movement, limited visibility, occlusions, and multi-object interactions within InterVLA. Additionally, we emphasize the dataset’s potential applications in tasks such as sparse-view 4D scene reconstruction, hand-object interaction, and motion reconstruction from sparse signals. Our contributions can be summarized as:
\begin{itemize}
\item We collect the first large-scale human-object-human interaction dataset called InterVLA with diverse generalist interaction categories and egocentric perspectives.
\item Our proposed benchmark with thorough analysis on egocentric human motion estimation, interaction synthesis and interaction prediction will stimulate future works on building intelligent AI assistants. We will release all the datasets, code and models for further exploration.
\end{itemize}

% \vspace{-2mm}
\section{Related Work}
\label{sec:related_work}
% \vspace{-1mm}

\noindent \textbf{Egocentric Vision.} Increasing attention is attached to egocentric vision with datasets~\cite{grauman2022ego4d,grauman2024ego,wang2023holoassist,li2023ego,hao2024ego3dt,damen2022rescaling,damen2018scaling,fathi2012social,li2018eye,nakamura2017jointly,HOT3D}, spurred by applications like robotics, AR and VR. The unique dynamics and perspective of egocentric vision present new challenges and opportunities for various tasks, including wearer pose estimation~\cite{li2023ego,luo2021dynamics,wang2023scene,millerdurai2024eventego3d,wang2024egocentric,jiang2023egoposer,akada20243d}, activity recognition~\cite{kazakos2019epic,li2021ego,zhou2015temporal,grauman2024ego}, human-object interaction (HOI)~\cite{kwon2021h2o,cai2016understanding,damen2016you,nagarajan2019grounded,ma2024diff-ip2d}, robotic active perception~\cite{tirumala2024learning,xu2024humanvla,khatibi2020real,an2024rgbmanip}, and interactive assistants~\cite{wang2023holoassist}. InterVLA captures the egocentric data via two GoPro cameras showing in ~\cref{fig:data_capture}, together with accurate humans and objects motions obtained by the OptiTrack MoCap system.
Compared to most video datasets such as Ego-Exo4D~\cite{grauman2024ego}, InterVLA provides ground-truth 4D human and object motions captured by an optical MoCap system.
% poses and meshes of human and objects supported from an OptiTrack MoCap system and mesh reconstruction scheme.

\noindent \textbf{Human-Human Interactions.} Besides numerous single-human motion datasets~\cite{nturgbd120,humanact12,uestc,punnakkal2021babel,plappert2016kit,humanml3d,motion-x,wang2024quovadis,xu2024motionbank}, several human-human datasets~\cite{van2011umpm,sbu_kinect,ng2020you2me,chi3d,nturgbd120,expi,hi4d,interhuman,inter-x} have also been constructed for interaction synthesis~\cite{xu2023actformer,interhuman,inter-x,fan2025go}, human reaction generation~\cite{inter-x,xu2024regennet}.~\cite{he2024omnih2o,jiang2024harmon,prasad2022mild,prasad2024moveint,butepage2020imitating} also verifies that the learned interaction knowledge can be applied to human-robot interactions.
InterVLA is essentially a human-human interaction dataset composed by an instructor and an assistant interacting with objects.

\noindent \textbf{Human-Object Interactions.} Recent efforts have expanded the boundaries of HOI datasets from hands-interactions~\cite{brahmbhatt2020contactpose,hampali2020honnotate,hampali2022keypoint,jian2023affordpose,liu2024taco,liu2022hoi4d} and full-body interactions~\cite{fan2023arctic,huang2022intercap,taheri2020grab,xu2021d3dhoi} with single object to full-body interactions with multiple objects~\cite{mandery2016unifying,lv2025himo}. In contrast to previous datasets focusing on individual HOI episode without context, InterVLA captures a series of coherent and consecutive HOI episodes for each GPT-generated script.

\noindent \textbf{Human-Scene Interactions.} Human-scene interactions incorporate comprehensive aspects ranging from navigation and collision avoidance to interaction with objects in the scene with various applications like embodied AI and VR. Real-world datasets~\cite{monszpart2019imapper,savva2016pigraphs,hassan2019resolving,guzov2024interaction,hassan2021stochastic,bhatnagar2022behave,jiang2023full,zhang2022couch} face limitations such as static scenes~\cite{guzov2024interaction,hassan2021stochastic}, single objects~\cite{bhatnagar2022behave,jiang2023full,zhang2022couch}, or noisy 3D pose estimated from image~\cite{monszpart2019imapper,savva2016pigraphs,hassan2019resolving}. Synthetic datasets ~\cite{jiang2024scaling,araujo2023circle,black2023bedlam,cao2020long} resolve these issues yet suffer from appearance reality and physical plausibility. We build simple scenes with random furniture arrangements, offering comprehensive navigation and interaction data in real-world dynamic scenes. 

\noindent \textbf{Human-Object-Human Interactions.} HOH interactions are common in cooperative tasks and settings. Several datasets  ~\cite{cini2019choice,wiederhold2024hoh,kshirsagar2023dataset,carfi2019multi,faibish2022human} focusing on handover are limited by fixed scene and body position settings, lack of egocentric RGB data, and absence of either natural human appearance or 3D human pose data.
While CORE4D~\cite{zhang2024core4d} provides egocentric HOH interaction data, it focuses exclusively on a single interaction type of two person collaboratively rearranging objects.
InterVLA addresses those limitations to involve diverse HOH interaction types, flexible scene settings and multi-object interactions.

\begin{figure*}[t]
  % \vspace{-2mm}
  \centering
  \includegraphics[width=1.0\linewidth]{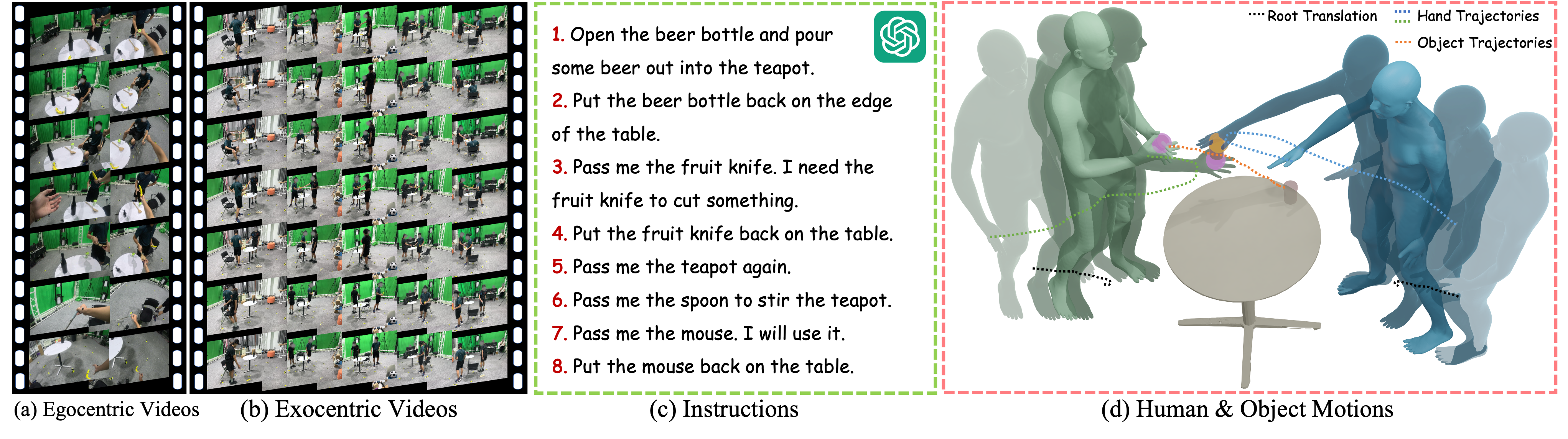}
  % \vspace{-2mm}
  \caption{\textbf{Components of the InterVLA dataset.} For the \textbf{vision} modality, we capture (a) two egocentric and (b) five exocentric RGB videos; For the \textbf{language} modality, we comprehensively supply the (c) GPT-generated commands; For the \textbf{action} modality, we provide the high-precision (d) human and object motions during the interactions.}
  \label{fig:vla_sample}
  % \vspace{-5mm}
\end{figure*}

\noindent \textbf{Vision-Language-Action Models.} VLA models integrate vision and language as multi-modal inputs and generate actions for agents to accomplish tasks. The last few years have witnessed many great works~\cite{kim2024openvla,brohan2023rt,zhen20243d,brohan2022rt,driess2023palm,o2023open,huang2023voxposer,ma2024survey} propelling the advancement of VLA models with the support of large pre-trained models. HumanVLA~\cite{xu2024humanvla} pioneeringly simulated humanoid VLA using simulator-style images as vision input. We anticipate InterVLA with real-world egocentric RGB data, language instructions, human response motion and object trajectories, could accelerate the practical humanoid VLA in real-world applications.

\section{The InterVLA dataset}
\label{sec:dataset}

% General insight of the dataset
\subsection{Overview}
InterVLA is a large-scale human-object-human interaction dataset collected in a vision-language-action scheme, which features an assistant providing diverse services to the instructor in daily scenarios. 
We believe that our two-person and multi-object setting integrates several specialist human-centric interactions and will facilitate further research on robot-centric interactions. 
Besides, our InterVLA dataset also emphasizes the utility of egocentric perception and the assistant's action based on it.
As depicted in~\cref{fig:teaser}, each scene comprises randomly arranged furniture with two persons interacting with several daily objects. 
The assistant performs a series of consecutive actions following the language commands of the instructor, such as \textit{``Give me the mug on the table''}.
We provide multi-view exocentric (third-person) viewpoints and two egocentric (first-person) perspectives from the assistant. The human and object motions are obtained by the optical motion capture (MoCap) system.
Next, we will describe the vision-language-action data collection scheme and pipeline in~\cref{sec:vla_capture} and dataset post-processing, components, and statistics in~\cref{sec:data_component}.

% Dataset collection process, divided by vision, language and action parts
\subsection{Vision-Language-Action Capturing}
\label{sec:vla_capture}

The concept of vision-language-action (VLA) emerges with the rise of instruction-following robotics. 
Inspired by it, we embed the manual-assisted task into the VLA framework where the assistant performs diverse services to the instructor such as picking up, retrieval, handover, and rearrangements of multiple objects.
The hybrid RGB-MoCap capturing system of InterVLA is elaborately illustrated in~\cref{fig:data_capture}.

\noindent \textbf{Capturing Pipeline.}
We recruit 46 participants to form 27 unique instructor-assistant pairs for data collection. 
For the object and scene setup, we primarily select 50 common household objects of various sizes, including small objects such as fruit, mug, knife, and large objects such as suitcases, floor hangers, and besom. Note that we adopt real objects rather than 3D printed objects as~\cite{taheri2020grab,lv2025himo} for the fidelity of the RGB modality. The details of the object list are provided in the supplementary material.
To align with real-world intelligent robotic assistance and enrich the interaction categories, we meticulously develop the following data-capturing pipeline.
We first randomly arrange some furniture of tables or chairs in the MoCap venue with several operable objects positioned in the scene. During collection, the instructor first communicates with the assistant using verbal commands along with complementary body gestures, while the assistant should interpret the instructor's intention and react appropriately. For each recording, a sequence of atomic interactions is performed to preserve long-duration interactions and maintain continuity.

% Five exo-centric iphones and two ego-centric gopros
\noindent \textbf{Vision.}
Two egocentric GoPro cameras are mounted tightly on the forehead and chest of the \textit{assistant}, respectively, to capture first-person RGB videos with a high resolution of 5312$\times$2988 at 30 fps. The camera intrinsics are obtained following~\cite{easymocap}. The camera positions and orientations are carefully adjusted based on the height of the participants to maximize the valid shooting area of the scene, the other person, and the interactions.
Besides, we also integrate five well-calibrated RGB cameras to compensate for the multi-view exocentric viewpoints with a resolution of 1920$\times$1080 at 30 fps. The egocentric and exocentric cameras are all temporally synchronized by millisecond-level timestamps.
To protect the privacy of participants, we mask the faces of all the exocentric and egocentric RGB videos with~\cite{haarcascades}.

\noindent \textbf{Language.}
Commands of the instructor serve as the starting point, trigger, and bridge for the following interactions. Given the household objects and furniture library, we employ large language models, \ieno, ChatGPT~\cite{gpt3.5} to select the scenes and objects setup, determine their placement, and then generate a script of instructor-assistant interaction sequences within the scene based on the object affordance. 
With the majority of interactions focusing on human-object-human interactions such as handover and collaborative rearrangement, we also include some pure human-human interactions such as supporting or massaging. Furthermore, we encourage multi-object interactions with object-object interactions such as \textit{``Slicing the apple with a knife''} and simultaneous manipulation such as \textit{``Tidying up the objects on the table''}. Additionally, moving around and navigating within the scene are also supported.
Ultimately, we produce 100 scripts, each involving an average of 2-3 furniture, 5 household objects, and 8 consecutive atomic commands.
Each script is manually reviewed to ensure validity. % and coherence.

\noindent \textbf{Action.}
For robotic arms, \textit{action} can be defined as the rotations and translations of the robot joints. Similarly, motion is a compact representation for modeling human actions and object movements.
To acquire high-quality human and object motions, we establish a MoCap system of 8.5$\times$5.4 meters with 20 infrared cameras.
For the human motions, we discard the tight MoCap suits but instead directly attach the reflective markers on the skin or clothes surface by strong medical glue as~\cite{ionescu2013human3,fan2023arctic} to preserve the fidelity of the RGB modality. The relative displacement between skin and clothing is eliminated by glue to ensure robust and precise MoCap results. 
The objects are treated as rigid bodies with at least four reflective markers placed on the surfaces for optical tracking. 
In our setting, we adopt the 12.5mm diameter reflective spheres for both humans and objects to achieve the best tracking results.
All the assets are well-created and calibrated before recording.
We also equip the MoCap system with timecodes for the temporal alignment with the ego-exo RGB videos.

% How to postprocess the dataset, statistics of the dataset
% 1. SMPL, 2. Hand, 3. Clips
\subsection{Dataset Components}
\label{sec:data_component}

\noindent \textbf{Human Parametric Model.}
SMPL parametric model~\cite{smpl} is widely adopted in human-centric interaction datasets, which formulates the human mesh as the body pose $\theta\in \mathbb{R}^{23\times3}$, global orientation $q_i\in \mathbb{R}^{3}$, root translation $\gamma_i\in \mathbb{R}^{3}$ and the body shape parameters $\beta\in \mathbb{R}^{10}$, which are determined based on the height, weight and gender of the participant as~\cite{virtual_caliper,inter-x,lv2025himo}.
We fit the BVH-format human skeleton captured from the MoCap data to the SMPL parameters with the following optimization objective as:
\begin{equation}
    \mathcal{L} = \lambda_j \mathcal{L}_j + \lambda_s \mathcal{L}_s + \lambda_{reg} \mathcal{L}_{reg},
\end{equation}
where $\mathcal{L}_j$ measures the difference between the raw MoCap joint position and the optimized result, $\mathcal{L}_s$ smooths the inter-frame motion transitions and mitigates pose jittering, $\mathcal{L}_{reg}$ regularizes the optimized poses from deviating and $\lambda_j=1$, $\lambda_s=0.1$, $\lambda_{reg}=0.01$ are loss weights. We give further details of each optimization term in supplementary.

\noindent \textbf{Object Meshes and Tracking.}
To render accurate human-object interactions, we scan all 50 objects and obtain the precise object surface geometries with a KSCAN Magic Scanner. Each object is then attached by more than 3 reflective markers and tracked by the optical MoCap system. 
We further scan the objects together with the attached markers and align the new scans with the previous results to eliminate the offsets between the two centroids. The object motion can be represented as the translation $t^o\in \mathbb{R}^3$ and rotation $r^o\in \mathbb{R}^6$ of the 6D rotation representation~\cite{6d_rot}.

\noindent \textbf{Alignment and Segmentation.}
We standardize all the video data to a resolution of 1920$\times$1080 at 30 fps and downsample the MoCap data to the same fps for consistency. The exocentric videos, egocentric videos and motion data are well synchronized with millisecond-level timestamps.
To facilitate training for downstream tasks, we render the interaction results and then manually split the long-duration script into short clips by the atomic commands while retaining the temporal continuity across the clips. The RGB videos are also segmented in the same way.

\begin{figure*}[t]
  % \vspace{-2mm}
  \centering
  \includegraphics[width=1.0\linewidth]{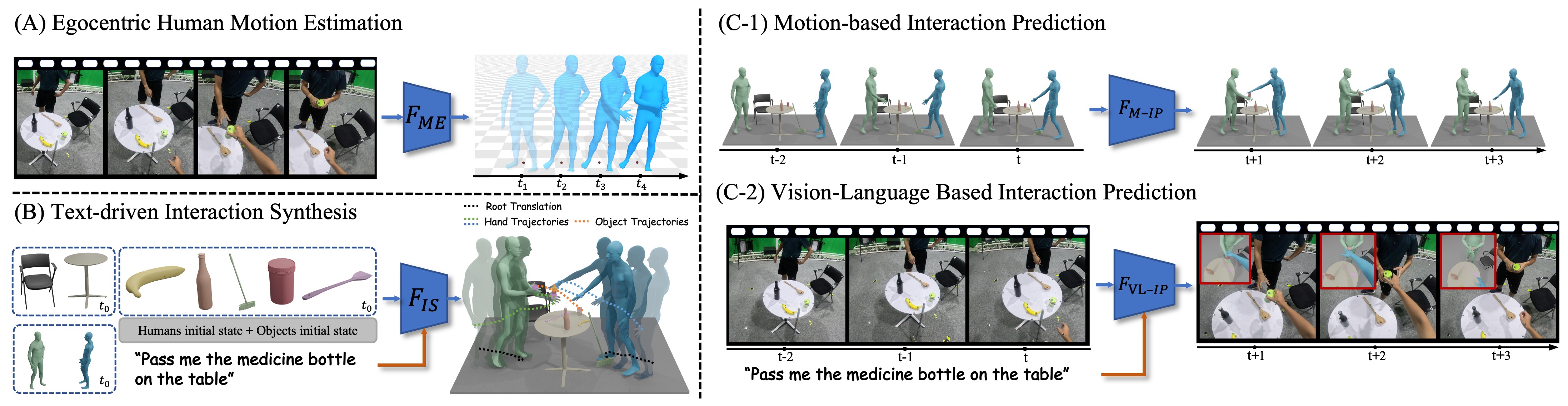}
  % \fbox{\rule{0pt}{2in} \rule{.9\linewidth}{0pt}}
  % \vspace{-2mm}
  \caption{\textbf{Task Formulation of InterVLA.} We establish multiple downstream tasks on egocentric human motion estimation, text-driven interaction synthesis, motion-based interaction prediction and vision-language based interaction prediction. All these benchmarks showcase the great challenges of our InterVLA dataset and will benefit practical, intelligent AI assistants.}
  \label{fig:tasks}
  % \vspace{-5mm}
\end{figure*}

\noindent \textbf{InterVLA Statistics.}
As aforementioned, we recruit 47 participants forming 27 unique instructor-assistant pairs for the data collection, spanning 519 valid long-duration interaction recordings with different GPT-generated scripts and scene arrangements. We ensure that each object appears in at least five scenarios. After the alignment, segmentation and thorough manual refinement of the motion data, we obtain 3,906 interaction sequences with 11.4 hours and 1.2M frames of four-tuple data composed by the egocentric videos, exocentric videos, instructions and human/object motions. We showcase an example of the dataset in~\cref{fig:vla_sample} for a better illustration and more video examples will be provided in the supplementary materials.

\section{Tasks, Benchmarks and Experiments}

We define new tasks and benchmarks of egocentric human motion estimation, interaction synthesis and interaction prediction as presented in~\cref{fig:tasks} and potential downstream tasks based on InterVLA, oriented towards general-purpose AI agents. We also provide the evaluation results for the proposed benchmarks with detailed analysis.

\subsection{Preliminary Formulation}
% $[V_e, V, X, O, \bm{O}, \mathbf{O}, \mathcal{L}, \bm{X}, \mathbb{O}]$

Egocentric videos exhibit the active perspective that shows the superiority of capturing the details of the close-by interactions and the first-person intentions. However, they also suffer from the constrained camera field of view without global perception, rapid viewpoint changes of objects exiting and re-entering the field of the frame.

We formulate our multimodal data as $[\bm{V}, \bm{T}, \bm{H}, \bm{O}, \bm{l}]$ for each sequence, where $\bm{V}=\{\bm{v}_h, \bm{v}_c\}$ are the \textbf{h}ead-mounted and \textbf{c}hest-mounted egocentric videos respectively, $\bm{T}=\{\bm{t}_i\}_{i=0}^{N_t}$ are the exocentric videos and $N_t=5$ indicates the video number. The human motions are denoted as $\bm{H}=\{\bm{h}_I, \bm{h}_A\}$ for the \textbf{I}nstructor and \textbf{A}ssistant respectively, where $\bm{h}_I$/$\bm{h}_A$ can be detailed as $\{\theta\in \mathbb{R}^{23\times6}, q\in \mathbb{R}^{6}, \gamma\in \mathbb{R}^{3}, \beta\in \mathbb{R}^{10}\}$ for the body pose, global orientation, root translation and the body shape parameters, respectively. The object motions can be represented as $\bm{O}=\{\bm{o}_i\}_{i=0}^{N_o}$, where $\bm{o}_i = \{r_i^o\in \mathbb{R}^{6}, t_i^o\in \mathbb{R}^{3}\}$ for the rotation and translation of the object, respectively. $N_o$ means the number of objects. The geometries of all involved objects are precisely scanned denoted as $\bm{G}=\{\bm{g}_i\}_{i=0}^{N_o}$. Here we adopt the 6D rotational representation~\cite{6d_rot} for humans and objects as in previous works~\cite{humanml3d,actor}. $\bm{l}$ refers to the language commands.
We split the dataset into training, testing and validation sets with the ratio of 0.8, 0.15 and 0.05 for all downstream tasks. 
Note that we adopt \textbf{only the head-mounted camera} $\bm{v}_h$ for all the video-based experiments.

\subsection{Egocentric Human Motion Estimation}

\noindent \textbf{Task Formulation.}
Egocentric perception and comprehension of the instructor's intention are fundamental as the first step of AI assistants. As depicted in~\cref{fig:tasks}~(A), we aim to reconstruct the world-grounded human motion sequences given the egocentric RGB videos as
\begin{equation}
  F_{ME}(\bm{v}_h) \mapsto \hat{\bm{h}_I},
  \label{eq:me}
\end{equation}
where $\hat{\bm{h}_I}$ denotes the estimated instructor motion.
Note that it is quite challenging to maintain accurate body pose estimation and keep the consistent global coordinate system.
The rapid movement of egocentric cameras, occlusion and limited visibility raise substantial challenges for this task.

\noindent \textbf{Experiment Settings.}
We evaluate four state-of-the-art global human motion estimation methods, TRACE~\cite{sun2023trace}, GLAMR~\cite{yuan2022glamr}, TRAM~\cite{wang2025tram}, and WHAM~\cite{shin2024wham} on all the head-mounted camera sequences of the InterVLA dataset for an intuitive understanding.

\noindent \textbf{Evaluation Metrics.}
Following previous works~\cite{shin2024wham,wang2025tram}, We compute Mean Per Joint Position Error (MPJPE), Procrustes-aligned MPJPE (PA-MPJPE) and Per Vertex Error (PVE) to evaluate the 3D human pose and body shape estimation performance, and Acceleration error (Accel) for the inter-frame motion smoothness. More specifically, MPJPE calculates the average of the Euclidean distances between the predicted and ground truth joint positions. PA-MPJPE is the MPJPE calculated after Procrustes analysis that aligns the predicted poses to the ground truth through translation, rotation, and scaling. PVE measures the average distance between predicted and ground truth positions of the 6,890 vertices derived from the SMPL parametric model. Accel calculates the average difference in acceleration between the predicted and ground-truth coordinates.

\begin{figure*}[t]
  % \vspace{-2mm}
  \centering
  \includegraphics[width=1.0\linewidth]{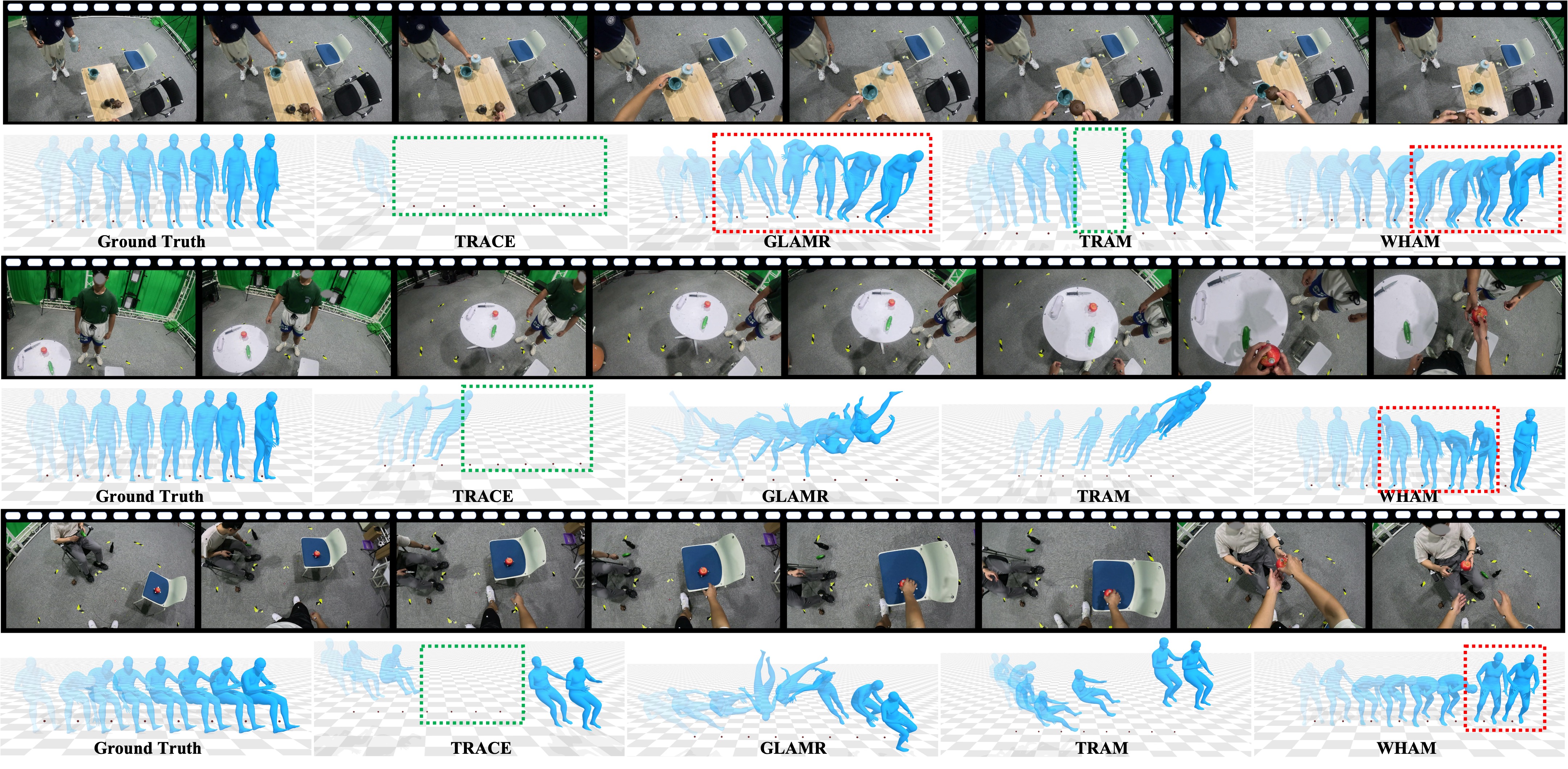}
  % \fbox{\rule{0pt}{2in} \rule{.9\linewidth}{0pt}}
  \caption{\textbf{Visualization result comparison} of the egocentric world-grounded human motion estimation results on the InterVLA dataset. For each sequence, we provide the original RGB frames along with the ground truth motion obtained by the MoCap system and four baseline models. Higher opacity indicates later frames of the sequence. Please zoom in for a more detailed view.}
  \label{fig:ego_hmr}
  
\end{figure*}

\noindent \textbf{Results and Analysis.}
We present the visualization results comparison in~\cref{fig:ego_hmr} with the raw egocentric frames, the ground truth motion captured by the MoCap system and the baseline results. Note that we manually add a horizontal offset to avoid the motion entanglement, yet we place the \textit{anchor points} on the floor to refer the global translation of the motion. The results show that even the best-performing algorithms obtain unsatisfactory results. For the three samples, the assistant turns head to locate scene objects slightly (first-row) or rapidly (third-row), causing the instructor to move out of the camera's field of view. The red red dashed boxes show that GLAMR and WHAM fail to track the correct global orientation and keep the correct coordinate system. The green dashed boxes show that TRACE misses many frames.
Besides, we also provide the quantitative results comparisons in~\cref{tab:ego_hmr}, where WHAM~\cite{shin2024wham} show remarkable superiority over other methods across all metrics except PA-MPJPE. However, there remains a significant gap between its performance and the ground truth. This discrepancy can be attributed to several challenges inherent in our dataset, such as occlusions, rapid camera motion movements, the incomplete capture of human bodies, and frequent occurrences of subjects entering and exiting the frame (re-entering).

\begin{table}[t]
  % \vspace{-2mm}
  \centering
  \begin{adjustbox}{width=0.8\columnwidth, center}
  \begin{tabular}{@{}lcccc@{}}
    \toprule
    Method & PA-MPJPE$\downarrow$ & MPJPE$\downarrow$ & PVE$\downarrow$ & Accel$\downarrow$ \\
    \midrule
    TRACE~\cite{sun2023trace} & \textbf{91.2} & 720.1 & 761.3 & 31.0 \\
    GLAMR~\cite{yuan2022glamr} & 134.8 & 589.9 & 596.7 & 24.9 \\
    TRAM~\cite{wang2025tram} & 102.7 & 684.1 & 718.7 & 27.5 \\
    WHAM~\cite{shin2024wham} & 103.2 & \textbf{333.6} & \textbf{359.7} & \textbf{8.7} \\
    \bottomrule
  \end{tabular}
  \end{adjustbox}
  \caption{\textbf{Quantitative results} of egocentric global human motion estimation on the InterVLA dataset. \textbf{Bold} for the best results.}
  \label{tab:ego_hmr}
\end{table}

\subsection{Interaction Synthesis}

\noindent \textbf{Task Formulation.}
Following existing text-driven HOI synthesis methods~\cite{hoi_diff,lv2025himo,cg_hoi,HOI-M3}, we define the task as multiple humans and objects motion generation as shown in~\cref{fig:tasks}~(B). We formulate this task as
\begin{equation}
  F_{IS}(\bm{H}^0, \bm{O}^0, \bm{G}, \bm{l}) \mapsto \{\hat{\bm{H}}, \hat{\bm{O}}\},
  \label{eq:is}
\end{equation}
where $\bm{H}^0$ and $\bm{O}^0$ represent the initial pose of the humans and objects respectively, and $\hat{\bm{H}}$ and $\hat{\bm{O}}$ are the generated human and object motions.
Compared with previous HOI datasets such as BEHAVE~\cite{bhatnagar2022behave}, our InterVLA dataset contains multiple humans and objects and multi-object manipulations, which introduces greater challenges.

\noindent \textbf{Experiment Settings.} Text-driven human motion generation methods MDM~\cite{mdm}, priorMDM~\cite{priormdm} and HIMO~\cite{lv2025himo} are re-implemented to support the condition input of object meshes and the initial states of two persons and multiple objects. Further details regarding these methods can be found in the supplementary materials.

\begin{table}[t]
  \centering
  \resizebox{0.48\textwidth}{!}{
  \begin{tabular}{@{}lccccc@{}}
    \toprule
    Methods  & R Precision (Top 3) $\uparrow$ & FID $\downarrow$ & MM Dist $\downarrow$ & Diversity $\rightarrow$ & MModality $\uparrow$\\
    \midrule
      Real & $0.7592^{\pm0.0026}$ & $0.0203^{\pm0.0018}$ & $3.8718^{\pm0.0064}$ & $9.0164^{\pm0.0831}$ & $-$ \\
      \midrule
      MDM~\cite{mdm}& $0.4897^{\pm0.0067}$ & ${2.8039^{\pm0.0727}}$ & $5.4879^{\pm0.0212}$ &  $7.7260^{\pm0.0633}$ &  $1.9888^{\pm0.0694}$ \\
      priorMDM~\cite{priormdm} & $0.5250^{\pm0.0068}$ & $6.2766^{\pm0.0777}$ & $5.5129^{\pm0.0188}$ & $8.9414^{\pm0.0888}$ & $\mathbf{2.1227^{\pm0.0993}}$\\
      HIMO~\cite{lv2025himo} & $\mathbf{0.5707^{\pm0.0029}}$ & $\mathbf{0.6805^{\pm0.0136}}$ & $\mathbf{4.9609^{\pm0.0162}}$ & $\mathbf{8.9849^{\pm0.0554}}$ & $1.1478^{\pm0.0658}$\\
    \bottomrule
  \end{tabular}}
  \caption{\textbf{Quantitative results} of human-object-human interaction synthesis on the InterVLA dataset, where $\pm$ indicates 95\% confidence interval and $\rightarrow$ means the closer the better. \textbf{Bold} highlights the best results.}
  \label{tab:hoi_synthesis}
\end{table}

\noindent \textbf{Evaluation Metrics.}
We train the text feature extractor and human-object motion feature extractor first via contrastive learning as~\cite{humanml3d,lv2025himo}. The generation quality is evaluated by the following metrics: R Precision to evaluate the top-3 accuracy in retrieving the ground-truth description, Frechet Inception Distance (FID)~\cite{heusel2017gans} to measure the latent space divergence between authentic and synthetic samples, MultiModal distance (MM Dist) to determine the latent space distance between generated motions and input texts, Diversity to gauge the variance within the latent space, multimodality (MModality) to quantify the diversity of outputs generated from the same textual input.

\noindent \textbf{Results and Analysis.} The quantitative results presented in \cref{tab:hoi_synthesis} demonstrate that HIMO~\cite{lv2025himo} surpasses other methods across all metrics except for MModality, with a particularly impressive performance on FID.
All these methods achieve a higher FID than the real interaction data, which shows that there remains ample opportunity for future endeavors to enhance the naturalness of the generated interaction results.

\subsection{Interaction Prediction}

\noindent \textbf{Task Formulation.}
We propose two types of interaction prediction tasks of motion-based and vision-language based as demonstrated in~\cref{fig:tasks}~(C1) and~(C2). For the motion-based interaction prediction, the model predicts the subsequent HOI sequences for the following frames given the adjacent past few frames of HOI sequences as 
\begin{equation}
  F_{M\text{-}IP}(\bm{H}^{t_I:t-1}, \bm{O}^{t_I:t-1}, \bm{G}) \mapsto \{\hat{\bm{H}}^{t:t_E}, \hat{\bm{O}}^{t:t_E}\},
  \label{eq:mip}
\end{equation}
where $t_I$ ($t_E$) denotes the initial (ending) frame of the sequence, $\hat{\bm{H}}^{t:t_E}$ and $\hat{\bm{O}}^{t:t_E}$ are the predicted future motions. In the experiments, we set $t_I=0$, $t=15$ and $t_E=30$ to predict the poses in the subsequent 15 frames given the previous 15 frames.
While for the vision-language based interaction prediction, the model manages to anticipate the future human motions from the historical egocentric videos and the verbal instruction as
\begin{equation}
  F_{VL\text{-}IP}({\bm{v}_h}^{t_I:t-1}, \bm{h}_A^{t_I:t-1}, \bm{l}) \mapsto \{\hat{\bm{h}_A}^{t:t_E}\},
  \label{eq:mip}
\end{equation}
where $\hat{\bm{h}_A}^{t:t_E}$ means the predicted motion of the assistant.

\noindent \textbf{Experiment Settings.}
We evaluate three state-of-the-art methods CAHMP~\cite{corona2020context}, MDM~\cite{mdm} and InterDiff~\cite{xu2023interdiff} for motion-based interaction prediction. For vision-language guided interaction prediction, we re-implement the existing hand trajectory prediction models FHOI~\cite{liu2020forecasting}, OCT~\cite{liu2022joint}, and USST~\cite{bao2023uncertainty} with integrated language embeddings to predict the motion of the assistant.

\begin{table}[t]
  % \vspace{-2mm}
  \centering
  \begin{adjustbox}{width=\columnwidth, center}
  \begin{tabular}{@{}lccccc@{}}
    \toprule
    \multirow{2}{*}{Methods} & Human & \multicolumn{2}{c}{Object} & \multicolumn{2}{c}{Contact} \\
    \cmidrule(lr){2-6}& $J_e$(mm, $\downarrow$) & $T_e$(mm, $\downarrow$) & $R_e$($^\circ$, $\downarrow$) & $C_{acc}$(\%, $\uparrow$) & $P_r$(\%, $\downarrow$) \\
    \midrule
    MDM~\cite{mdm} & 175.3 ($\pm$ 0.8) & 140.2 ($\pm$0.7) & 11.0 ($\pm$0.2) & 85.5 ($\pm$0.3) & 0.4 ($\pm$0.0) \\
    InterDiff~\cite{xu2023interdiff} & 175.3 ($\pm$ 0.8) & 138.7 ($\pm$0.6) & 10.8 ($\pm$0.1) & 86.0 ($\pm$0.2) & 0.4 ($\pm$0.0)  \\
    CAHMP~\cite{corona2020context} & \textbf{172.5 ($\pm$ 0.4)} & \textbf{115.6 ($\pm$0.5)} & \textbf{9.5 ($\pm$0.1)} & - & - \\
    \bottomrule
  \end{tabular}
  \end{adjustbox}
  \caption{\textbf{Quantitative results} of motion-based interaction prediction on the InterVLA dataset.}
  \label{tab:action_predict}
\end{table}

\noindent \textbf{Evaluation Metrics.}
For motion-based interaction prediction, we follow \cite{zhang2024core4d} to apply the evaluation metrics of human joint position error ($J_e$) to measure the Mean Per Joint Position Error (MPJPE) for two individuals; object translation error ($T_e$) representing the average $\mathit{L}$2 differences between predicted and real object translations; object rotation error ($R_e$), indicating the average geodesic differences between predicted and actual object rotations; human-object contact accuracy ($C_{acc}$) to assess the average error rate in contact detection with a 5cm threshold to detect hand contacts; and penetration rate ($P_r$) calculating the percentage of object vertices penetrating human meshes.
For vision-language guided interaction prediction, we adopt the average displacement error as the average $\mathit{L}$2 distance between the predicted and ground truth trajectories, and the final displacement error as the $\mathit{L}$2 distance between the two final predicted and ground truth locations following FHOI~\cite{liu2020forecasting}.

\noindent \textbf{Results and Analysis.} We provide the quantitative results of three state-of-the-art models for motion-based interaction prediction in~\cref{tab:action_predict}. From the results, we can derive that CAHMP~\cite{corona2020context} achieves the best performance over the other baseline methods for both the human joint position error and the object translation and rotation error, thanks to the semantic-graph model to learn the relationship between human and context objects. 
From the quantitative comparisons of the vision-language based interaction prediction in~\cref{tab:vla_exp}, we can derive that USST~\cite{bao2023uncertainty} achieves the best performance for the two metrics due to the proposed uncertainty-aware state space Transformer.
However, we find that existing state-of-the-art methods have not achieved satisfactory performance on these two tasks. Significant errors remain in the predicted human motion, object translation \& rotation and human-object contact for these two tasks, which indicates that our dataset presents significant challenges for subsequent optimization.

\begin{table}[t]
\scriptsize
  \centering
  % \vspace{-3mm}
  \begin{tabular}{@{}lcc@{}}
    \toprule
    Method & Avg. Disp. Error $\downarrow$ &  Final Disp. Error $\downarrow$\\
    \midrule
    FHOI~\cite{liu2020forecasting} & 0.29 & 0.38\\
    OCT~\cite{liu2022joint} & 0.28 & 0.36\\
    USST~\cite{bao2023uncertainty} & \textbf{0.24} & \textbf{0.32}\\
    \bottomrule
  \end{tabular}
    % \vspace{-3mm}
  \caption{\textbf{Quantitative results} of vision-language based interaction prediction on the InterVLA dataset.}
  \label{tab:vla_exp}
  % \vspace{-3mm}
\end{table}

\subsection{Potential Downstream Tasks}

\noindent \textbf{Sparse-view 4D Scene Reconstruction.} Multi-view exocentric videos are synergistic with the egocentric viewpoint for global awareness of the scene. Existing works~\cite{zhang2024transplat,yu2024lm,xiong2023sparsegs,li2024dngaussian,wang2023sparsenerf} are dedicated to reconstructing static scenes from sparse-view images with 3D representations like meshes, neural radiance fields, and 3D Gaussian splatting while the majority of existing 4D reconstruction efforts~\cite{wu20244d,shih2024modeling,pumarola2021d,lin2024gaussian} are confined to dense perspective input. Our dataset comprising five exocentric-view videos, is anticipated to assist in downstream tasks of 4D scene reconstruction through additional supervisory signals, such as poses and meshes of people and objects.

\noindent \textbf{Hand-object Interaction Reconstruction.} 
Similar to~\cite{ye2022s,ye2023diffusion,fan2024hold,fan2025benchmarks} that jointly estimate the poses of both hands and the interacting objects, our InterVLA dataset with egocentric viewpoints and multi-object manipulation can also empower this task with substantial challenges.

\noindent \textbf{Motion Reconstruction from Sparse Signals.} This task aims to reconstruct one's own whole-body motion of the assistant given the sparse signals of the egocentric captured low body or arms as in previous works~\cite{jiang2022avatarposer,du2023agrol,yi2024estimating}.

\section{Conclusion}
\label{sec:conclusion}
In this paper, we introduce a large-scale egocentric human-object-human interaction dataset called InterVLA. By embedding the manual-assisted tasks into a vision-language-action scheme, we formulate \textit{vision} as the egocentric perspective, \textit{language} as the instructor's verbal commands and \textit{action} as the human and object motions, and demonstrate the indispensability of both generalist interaction knowledge and egocentric perception for building physical-world AI assistants. Through our extensive dataset and novel benchmarks for egocentric motion estimation, interaction synthesis and interaction prediction, we provide valuable tools that will drive further research and development in the field of real-world AI-assisted applications.

% \noindent \textbf{Limitations.}
% % % Hand
% % % Physical constraints
% % % Indoor, object categories, sequences limited
% While InterVLA is the first dataset designed for AI assistants where both the versatile human-centric interactions and egocentric perspective are considered, we highlight that some limitations remain. 1) First, InterVLA is limited to indoor scenarios with 50 daily objects involved. Extending our setting to outdoor settings or enriching the scenes are of great merit. Besides, indoor-captured dataset lack a certain level of realism, which is a common issue among indoor motion capture datasets. However, as the first dataset of its kind, we believe it holds significance for the broader human-robot interaction community. 2) Second, building InterVLA demands substantial time investment for attaching reflective markers, staging and changing the scenes and data processing. We strive to present InterVLA with $>$10 hours of high-quality interactive data, yet it is still insufficient for training large generalist interaction models. 3) Third, we discard the inertial gloves for capturing hand movements to preserve RGB realism. We apply several hand motion recovery models to InterVLA as illustrated before with extensive results and analysis.

\section*{Acknowledgements}
This work was supported in part by NSFC (62201342), Shanghai Municipal Science and Technology Major Project (2021SHZDZX0102), Grants of NSFC 62302246, ZJNSFC LQ23F010008, Ningbo 2023Z237 \& 2024Z284 \& 2024Z289 \& 2023CX050011 \& 2025Z038, and supported by High Performance Computing Center at Eastern Instituteof Technology and Ningbo Institute of Digital Twin. Authors would like to appreciate the Student Innovation Center of SJTU for providing GPUs.

{
    \small
    \bibliographystyle{ieeenat_fullname}
    \bibliography{main}
}

\input{suppl}

\end{document}

%% file: suppl.tex
\appendix

\begingroup
% \onecolumn 

\appendix
\twocolumn[
\begin{center}
\Large{\bf Perceiving and Acting in First-Person: \\A Dataset and Benchmark for Egocentric Human-Object-Human Interactions \\ **Appendix**}
\end{center}
]

\counterwithin{table}{section}
\counterwithin{figure}{section}

\section{Object Setting}

We list all 50 adopted objects of the InterVLA dataset in~\cref{tab:obj_cat}, which include 35 small objects and 15 large objects. 
Based on statistics, \textbf{41/100} scripts involves large object manipulations. 
For human-human interactions, InterVLA consists almost entirely of indirect human-human interactions through objects where the assistants need to comprehend the intention of the instructor before making responses. We also include some pure human-human interactions like ``support someone'' and ``wave''.
We provide more examples of large object manipulation and pure human-human interactions in~\cref{fig:large_obj} for better illustration.
In addition, we also provide the precise 3D object scans in the object\_mesh folder of \href{https://drive.google.com/drive/folders/1vojEFqxEFkhlHypDZha7EpalCGwNKePW?usp=sharing}{Google Drive} for your reference.

\begin{figure}[h]
  \centering
  % \vspace{-3mm}
  % \fbox{\rule{0pt}{0.5in} \rule{0.9\linewidth}{0pt}}
  \includegraphics[width=1.0\linewidth]{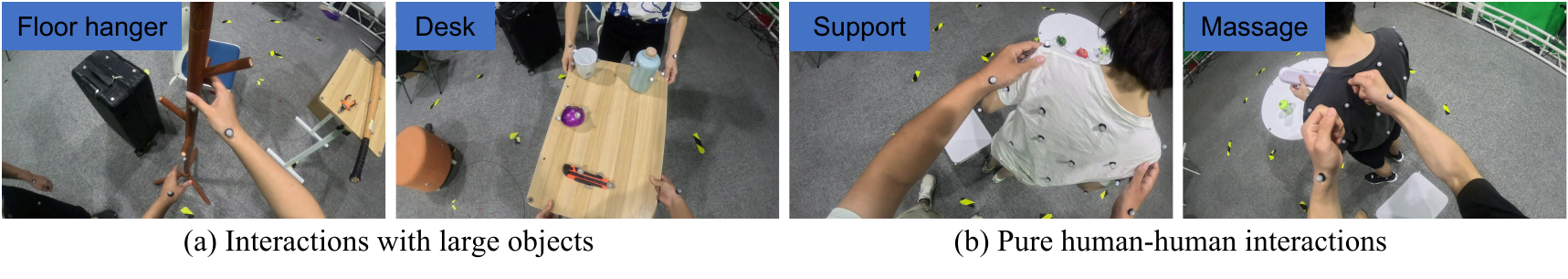}
  % \vspace{-8mm}
   \caption{\textbf{InterVLA samples.} More samples of large object manipulation and pure human-human interactions of InterVLA.}
   \label{fig:large_obj}
   \vspace{-5mm}
\end{figure}

\begin{table*}[!htbp]
  \centering
  \resizebox{1\textwidth}{!}{
  \begin{tabular}{|l|l|l|l|l|l|l|}
    \toprule
    01. Apple&02. Banana&03. Cucumber&04. Potato&05. Onion&06. Avocado&07. Orange \\ \midrule
    08. Liver bottle&09. Mid autumn box &10. Toothbrush box &11. Knife & 12. Spatula & 13. Ladle & 14. Spoon \\ \midrule
    15. Fork&16. Football &17. Mouse &18. Slipper left &19. Slipper right & 20. Remote control & 21. Wine bottle \\ \midrule
    22. Wine bottle black&23. Wine cylinder &24. Beer bottle &25. Teapot & 26. Tape & 27. Mug 1& 28. Mug 2 \\ \midrule
    29. Vacuum cup& 30. Hammer & 31. Piler & 32. Screwdriver & 33. Utility knife &34. Fruit knife&35. Rubbish bin \\ \midrule
    \textbf{36. Ukulele} &\textbf{37. Big box 1} &\textbf{38. Big box 2} &\textbf{39. Suitcase} &\textbf{40. Baseball bat} &\textbf{41. Besom} &\textbf{42. Dustpan} \\ \midrule
    \textbf{43. Floor hanger} &\textbf{44. Camera mount} &\textbf{45. Chair 1} &\textbf{46. Chair 2} &\textbf{47. Chair square} &\textbf{48. Sofa chair} &\textbf{49. Desk} \\ \midrule
    \textbf{50. Desk circle} & & & & & &\\ 
    \bottomrule
  \end{tabular}}
  \caption{\textbf{The objects setting of the InterVLA dataset.} The first 35 items are small objects, while the remaining 15 are large objects highlighted in \textbf{bold} font.}
  \label{tab:obj_cat}
  \vspace{-2mm}
\end{table*}

\section{SMPL Optimization}

Formally, the SMPL parameters consist of the body pose parameters $\theta \in \mathbb{R}^{N\times23\times 3}$, root translation $\gamma\in \mathbb{R}^{N\times3}$, global orientation $q\in \mathbb{R}^{N\times3}$, and the shape parameters $\beta\in \mathbb{R}^{N\times10}$, where $N$ indicates the number of frames. We initialize the shape of the participant $\beta$ based on their height and weight as~\cite{virtual_caliper}. Then, we optimize the SMPL parameters based on the Mocap data with the following optimization objective as:
\begin{equation}
    \mathcal{L} = \lambda_j \mathcal{L}_j + \lambda_s \mathcal{L}_s + \lambda_{reg} \mathcal{L}_{reg},
\end{equation}
where
\begin{equation}
    \mathcal{L}_j=\frac{1}{N}\sum\limits_{i=0}\limits^{N}\sum\limits_{j\in\mathcal{J}}\|\bm{J}_j^i(\mathbb{M}(\theta,\gamma,q)-\bm{g}_j^i\|_2^2
\end{equation}
aims to fit the SMPL joints to our captured skeleton data, where $\mathcal{J}$ denotes the joint set, $\mathbb{M}$ is the SMPL parametric model, $\bm{J}_j^i$ is the joint regressor function for joint $j$ at $i$-th frame, $\bm{g}$ is the Mocap skeleton data. A smoothing term 
\begin{equation}
    \mathcal{L}_{s}=\frac{1}{N-1}\sum\limits_{i=0}\limits^{N-1}\sum\limits_{j\in\mathcal{J}}\|\bm{J}_j^{i+1}-\bm{J}_j^{i}\|_2^2
\end{equation}
is applied to alleviate the pose jittering between frames. A regularization term 
\begin{equation}
    \mathcal{L}_{reg}=\|\theta\|_2^2
\end{equation}
is applied to constrain the pose parameters from deviating.

\section{Hand Pose Results}
We highlight the dexterous hand gestures such as manipulating objects and interacting with other individuals. However, attaching additional reflective markers on the hands fails to yield robust finger gestures empirically, and employing heavy inertial gloves significantly compromises the fidelity of RGB videos. 
To this end, we prioritize the natural RGB data of hand interactions and attach only three reflective markers to the hands to determine the rotation of the wrists. 
We notice that existing hand pose estimation algorithms~\cite{park2022handoccnet,yu2023acr,jiang2023probabilistic,pavlakos2024hamer,potamias2024wilor} demonstrate impressive accuracy and robustness even for in-the-wild hand images while other works~\cite{ye2022s,ye2023diffusion,fan2024hold,fan2025benchmarks} jointly estimate the poses of both hands and the interacting objects.
To this end, we apply the state-of-the-art hand pose estimation methods~\cite{potamias2024wilor} on our head-mounted egocentric videos as shown in~\cref{fig:wilor}, which yield robust estimated hand poses. 
% We provide an in-depth look at more hand pose examples, failure cases and the integration with the MoCap data to complete the high-quality hand-object interactions in the supplementary material.

\begin{figure}[!htbp]
  % \vspace{-2mm}
  \centering
  \includegraphics[width=1.0\linewidth]{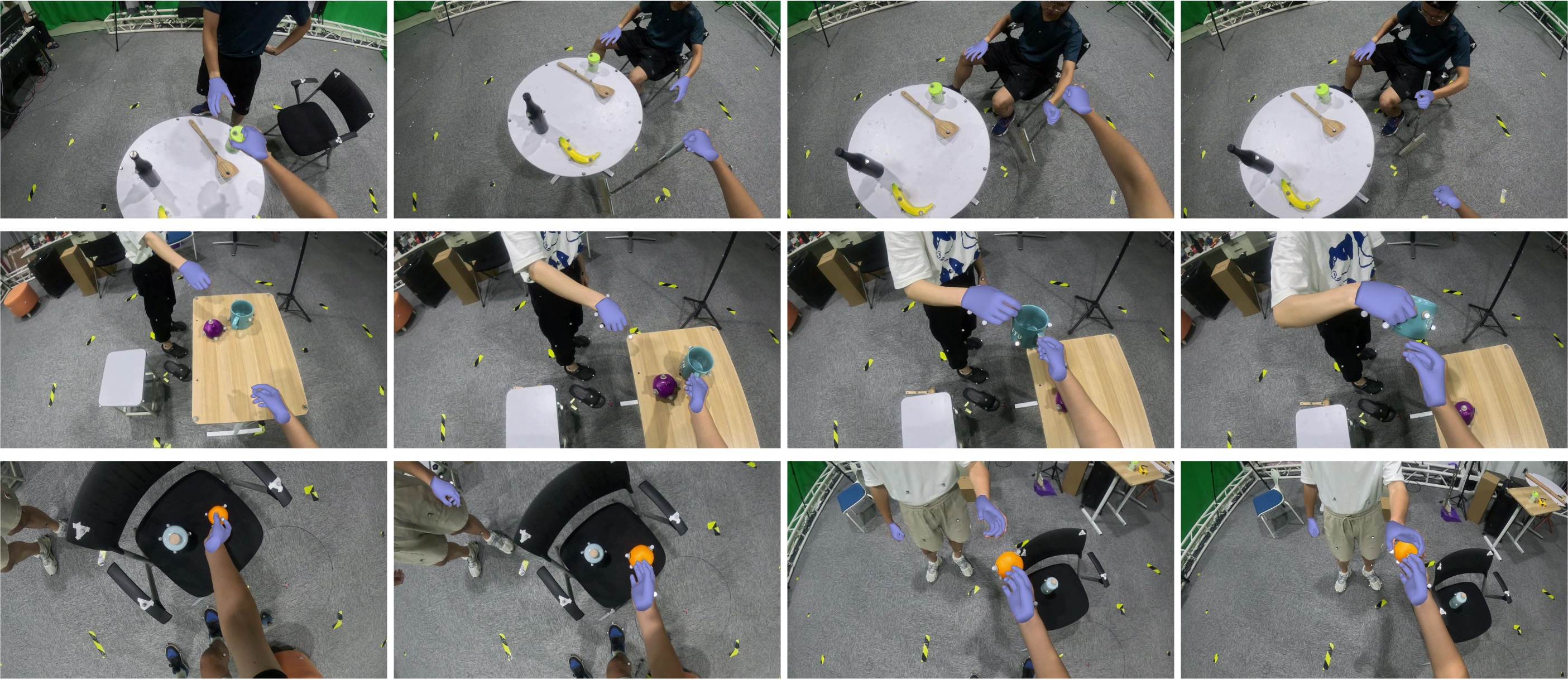}
  % \fbox{\rule{0pt}{2in} \rule{.9\linewidth}{0pt}}
  \caption{\textbf{Hand Pose Reconstruction}. Visualization results of the hand pose estimation results performed by WiLoR~\cite{potamias2024wilor} on the head-mounted egocentric videos of InterVLA.}
  \label{fig:wilor}
  \vspace{-2mm}
\end{figure}

We provide more visualization results of the failure cases of hand pose reconstruction of our InterVLA dataset by WiLoR~\cite{potamias2024wilor} in~\cref{fig:wilor_more}. We find that the state-of-the-art hand pose reconstruction method still fails to obtain smooth and accurate estimation results, which further validates the challenge of InterVLA. The failure parts are highlighted as red dashed boxes.

\begin{figure}[!htbp]
  % \vspace{-2mm}
  \centering
  \includegraphics[width=1.0\linewidth]{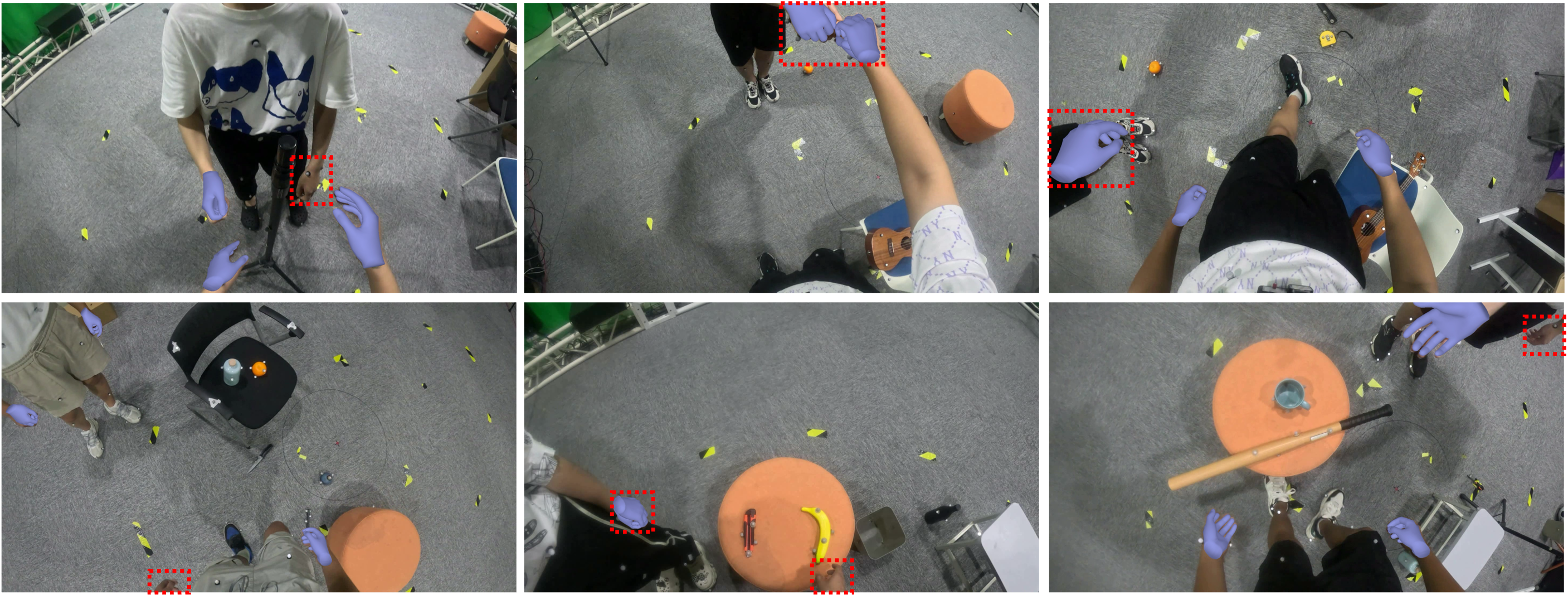}
  % \fbox{\rule{0pt}{2in} \rule{.9\linewidth}{0pt}}
  % \vspace{-2mm}
  \caption{\textbf{Failure Cases of Hand Pose Reconstruction}. We provide more results of the failure cases of the hand pose estimation results performed by WiLoR~\cite{potamias2024wilor} on the head-mounted egocentric videos of InterVLA.}
  \label{fig:wilor_more}
  % \vspace{-5mm}
  \vspace{-2mm}
\end{figure}

\section{Continuity between commands}
The continuity of operating one group of objects or one specific object is strictly guaranteed, such as ``give me the bottle'' $\to$ ``pure some wine into the bottle'' $\to$ ``put the bottle back on the table''. However, we don't emphasize the continuity among different objects with different functionalities. After the temporal segmentation process, all the atomic commands serve as independent VLA segments with complete semantics.

\section{More dataset examples}

We provide more dataset samples in the \href{https://drive.google.com/drive/folders/1vojEFqxEFkhlHypDZha7EpalCGwNKePW?usp=sharing}{Google Drive} together with the motion visualization tool implemented by ait-viewer~\cite{ait-viewer} (InterVLA\_Visualization\_Tool.zip). Please follow the instructions of the README to visualize the human and object motions of our dataset.

Besides, we also supplement the GPT-generated scripts, egocentric videos and exocentric videos in the \href{https://drive.google.com/drive/folders/1vojEFqxEFkhlHypDZha7EpalCGwNKePW?usp=sharing}{Google Drive} to provide a better demonstration of our dataset.

\section{Interaction Synthesis Settings}

\noindent{\textbf{MDM}}~\cite{mdm}.
We extend the original human motion generation model to human-object-human interaction generation, where the feature dimensions of the input and output are extended from $D_h$ to $D_h+D_o$, where $D_h$ is the dimension of human motion and $D_o$ denotes that of object motion. To embed the condition input of object geometries, we feed them into a linear layer and concatenate them with the initial poses of the objects. Then, all the conditions are concatenated with the noised input into the motion embedding.

\noindent{\textbf{PriorMDM}}~\cite{priormdm}.
The original PriorMDM~\cite{priormdm} is designed for two-person motion generation with two dual branches of MDM~\cite{mdm} and ComMDM to coordinate these two branches. We modify the two branches into a human motion branch and an object motion branch. Besides, we place the ComMDM module after the 4-th transformer layer of each branch to enable communication between the two branches. 

\noindent{\textbf{HIMO}}~\cite{lv2025himo}.
HIMO was designed for single-person interaction with multiple objects. We extend this method to two persons and up to seven objects. The object features are all concatenated together with the initial poses of these objects as the condition.

\section{Limitations.}
% % Hand
% % Physical constraints
% % Indoor, object categories, sequences limited
While InterVLA is the first dataset designed for AI assistants where both the versatile human-centric interactions and egocentric perspective are considered, we highlight that some limitations remain. 1) First, InterVLA is limited to indoor scenarios with 50 daily objects involved. Extending our setting to outdoor settings or enriching the scenes are of great merit. Besides, indoor-captured dataset lack a certain level of realism, which is a common issue among indoor motion capture datasets. However, as the first dataset of its kind, we believe it holds significance for the broader human-robot interaction community. 2) Second, building InterVLA demands substantial time investment for attaching reflective markers, staging and changing the scenes and data processing. We strive to present InterVLA with $>$10 hours of high-quality interactive data, yet it is still insufficient for training large generalist interaction models. 3) Third, we discard the inertial gloves for capturing hand movements to preserve RGB realism. We apply several hand motion recovery models to InterVLA as illustrated before with extensive results and analysis.